\documentclass[conference]{IEEEtran}
\IEEEoverridecommandlockouts

\usepackage{cite}
\usepackage{amsmath,amssymb,amsfonts}
\usepackage{algorithmic}
\usepackage{graphicx}
\usepackage{textcomp}
\usepackage{comment}
\usepackage{algorithm}
\usepackage{algorithmic}
\usepackage{hyperref}
\usepackage[table]{xcolor}
\usepackage[affil-it]{authblk} 

\def\BibTeX{{\rm B\kern-.05em{\sc i\kern-.025em b}\kern-.08em
    T\kern-.1667em\lower.7ex\hbox{E}\kern-.125emX}}
\begin{document}

\title{HiVeGen -- Hierarchical LLM-based Verilog Generation for Scalable Chip Design}

\author[1*]{Jinwei Tang}
\author[2*]{Jiayin Qin}
\author[1]{Kiran Thorat}
\author[2]{Chen Zhu-Tian}
\author[2]{Yu Cao}
\author[2]{Yang Katie Zhao}
\author[2]{Caiwen Ding}
\affil{*These authors contributed equally.}
\affil[1]{University of Connecticut - Storrs \hspace{0.5cm} \textsuperscript{2}University of Minnesota - Twin Cities}

\affil[1]{\texttt{tiw24001@uconn.edu}, \texttt{kiran\_gautam.thorat@uconn.edu}}
\affil[2]{\texttt{qin00162@umn.edu}, \texttt{ztchen@umn.edu}, \texttt{yucao@umn.edu}, \texttt{zhao1948@umn.edu}, \texttt{dingc@umn.edu}}

\renewcommand\Authsep{, } 
\renewcommand\Authands{, } 
\renewcommand\Authfont{\normalfont\normalsize} 
\renewcommand\Affilfont{\itshape\small} 

\maketitle 

\label{sec:abs}
\begin{abstract}
With Large Language Models (LLMs) recently demonstrating impressive proficiency in code generation, it is promising to extend their abilities to Hardware Description Language (HDL). However, LLMs tend to generate single HDL code blocks rather than hierarchical structures for hardware designs, leading to hallucinations, particularly in complex designs like Domain-Specific Accelerators (DSAs). 
To address this, we propose HiVeGen, a hierarchical LLM-based Verilog generation framework that decomposes generation tasks into LLM-manageable hierarchical submodules. HiVeGen further harnesses the advantages of such hierarchical structures by integrating automatic Design Space Exploration (DSE) into hierarchy-aware prompt generation, introducing weight-based retrieval to enhance code reuse, and enabling real-time human-computer interaction to lower error-correction cost, significantly improving the quality of generated designs.

\end{abstract}
\section{Introduction}
\label{sec:intro}

With the slowing pace of technology scaling and the explosive growth in artificial intelligence (AI)-driven applications, there is a growing demand for scalable chip designs that can effectively meet evolving application requirements while unlocking new capabilities. Domain-Specific Accelerator (DSA) chips exemplify this approach, as they are specifically customized to optimize particular applications, enabling significant improvements in power, performance, and area (PPA). 
Typically, DSAs, like other complex chip designs, employ hierarchical structures that decompose complex designs into manageable submodules. These hierarchical structures can be reprogrammed or tailored to meet various user/application requirements by adjusting their design configurations. However, designing DSAs remains an arduous and time-consuming venture. First, manually writing, debugging, and modifying hierarchical hardware description language (HDL) codes requires substantial design time and hardware design expertise, especially for complex designs. Furthermore, as design complexity grows, the design space expands, making design space exploration (DSE) a significant bottleneck that hinders both design efficiency and quality. 

Recent advancements in Large Language Models (LLMs) for natural language understanding and generation~\cite{openai2023gpt4} have inspired efforts to extend their ability to facilitate hardware chip designs. 
Prior works have demonstrated LLMs' potential in generating HDL code from natural language descriptions or high-level specifications\cite{thakur2023benchmarking,VHDL-Eval,MG-Verilog,Chipchat,Chipgpt,Chateda}. Some studies~\cite{GPT4AIGCHip,vungarala2024sa} have further explored the design space with the assistance of LLMs, through their ability to learn by imitation. 
However, the performance of LLM-generated hardware designs still needs improvements, especially for complex designs like DSAs. Through analyzing their outputs, we observe that LLMs tend to generate all code within \textit{one single block}. This results in excessively lengthy code, thereby leading to three critical issues: (1) \textit{Token length limit}: the length of code generation is constrained by the token limit of LLMs, making it challenging to successfully produce complex DSAs, let alone perform effective DSE. For example, LLMs may omit essential modules by replacing their implementations with comments to save tokens. (2) \textit{Code redundancy}: generating single code blocks disregards module reuse, leading to code redundancy and poor readability, making the design difficult to maintain and expand. (3) \textit{High error-correction cost}: 
as the single code block is generated at one time, users are unable to identify structural errors in real-time. 
However, continuously sending the user's prompt to LLMs for feedback -- the most straightforward approach -- is impractical due to computational demands and the unpredictable nature of LLM outputs~\cite{vaithilingam2022expectation}.
These limitations result in significant overhead when generating lengthy code for complex designs since any errors encountered require substantial effort to correct. 

Based on these observations, we propose HiVeGen, a hierarchical LLM-based Verilog generation framework that decomposes chip generation tasks into LLM-manageable hierarchical submodules and further reaps the hierarchical structures for DSE, code reuse, and low-cost error correction. The contributions of our proposed HiVeGen can be summarized as follows:

 \begin{figure*}[h!]
     \centering
\includegraphics[width=1.0\linewidth]{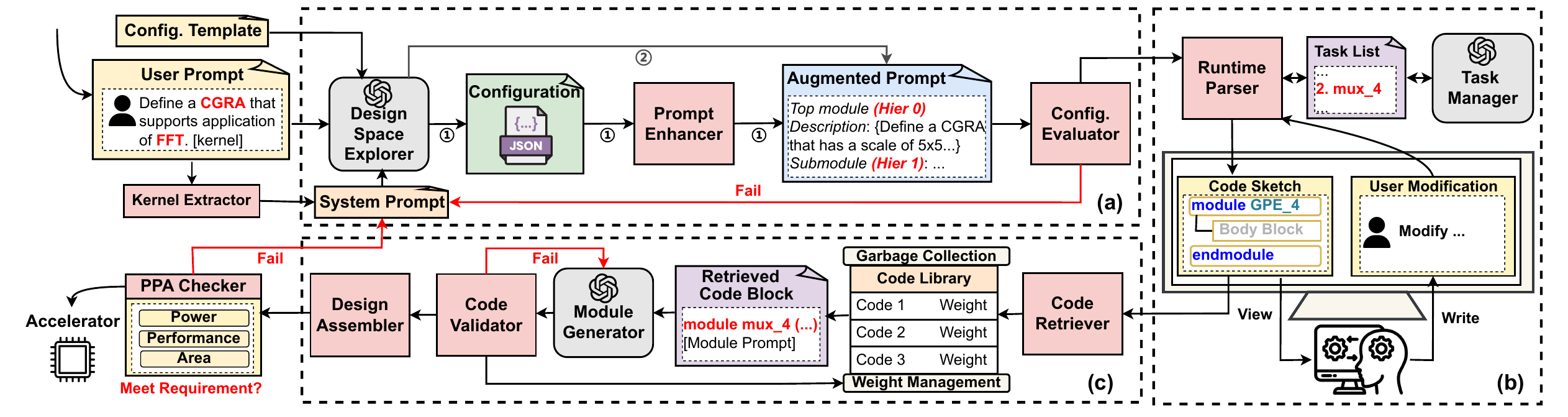} 
    \caption{The overview of our proposed HiVeGen framework with workflow. HiVeGen framework consists of (a) a Hierarchy-aware Prompt Generation Engine, (b) an On-the-fly Parsing Engine, and (c) a Weight-based Retrieving Engine. HiVeGen supports two design modes, each with its own workflow path: Path \textcircled{1}: generation path for DSA designs and Path \textcircled{2}: generation path for simple designs.}
    \label{fig1}
    \vspace{-6pt}
\end{figure*}

\begin{itemize}
\item To enable hierarchical structures, we propose a Hierarchy-Aware Prompt Generation Engine equipped with a Design Space Explorer to perform top-down hierarchical decomposition of designs. It iteratively optimizes the hierarchical configurations with PPA feedback from previous configuration rounds while maintaining alignment with application-specific constraints. 

\item Based on the hierarchical configuration, our framework employs a Weight-Based Retrieving Engine, which retrieves high-performance code blocks in the Code Library to reduce repeated generation while enhancing module reuse and quality. Innovatively, this engine assigns a weight parameter to each code block and dynamically updates it based on the likelihood of reuse, thereby reflecting the quality of the corresponding block.

\item We utilize an On-the-fly Parsing Engine that reduces the high correction cost by providing users with real-time interaction with the code structure while avoiding waiting for the access to the LLM-generated results. 
This enhances the generation accuracy as well as aligns the outputs with user expectations.

\item Compared to directly using LLMs, our framework achieves up to 45.24\% and 30.97\% runtime time savings and token savings, respectively, while enhancing generation accuracy. For DSA designs, the experiments demonstrate that our framework enables the generation of PPA-optimized accelerators with comprehensive DSE.
\end{itemize}

\section{Related Work}
\label{sec:rw}
Studies have long emphasized Electronic Deisign Automation (EDA) in hardware design to save effort in hardware accelerator design, which is primarily achieved by template-based customized toolchains or frameworks\cite{CGRA-ME}\cite{Gemmini}\cite{AHA}\cite{OpenCGRA}\cite{DeepBurning}\cite{DSAGEN}\cite{AutoDNN}\cite{RF-CGRA} developed by professional developers before the emerging of Large Language Model (LLM). However, these EDA tools tend to require users a high level of hardware expertise, for example, the complex domain-specific languages involved increase the design complexity and effort\cite{GPT4AIGCHip}.

To further improve hardware design speed as well as design accuracy, 
recent research on applying LLMs to chip design has produced promising results.  One of the pioneers, Chip-Chat~\cite{blocklove2023chip} delves into the intricacies of hardware design using LLMs, focusing on the task of register-transfer level (RTL) design by leveraging the capabilities of ChatGPT. 
RTLCoder shows how a lightweight LLM outperforms GPT-3.5 in RTL design generation using an open-source dataset, but similar to other works like Chip-Chat, it remains limited in terms of scaling to complex designs~\cite{liu2024rtlcoder, blocklove2023chip}. VerilogEval~\cite{liu2023verilogeval} assesses the performance of LLM in the realm of Verilog code generation for hardware design and verification. 
On the other hand, VGV~\cite{vgv2024}, extends LLM capabilities by integrating visual input with Verilog code generation, which increases the richness of the context but adds computational complexity, potentially slowing performance in larger designs. RTLLM~\cite{lu2023rtllm}
introduces a benchmarking framework consisting of 30 designs that are specifically aimed at enhancing the scalability of benchmark designs. Furthermore, it utilizes effective prompt engineering techniques to improve the generation quality.

Despite these improvements, the application of LLMs for complex VLSI Design, especially Domain-specific accelerators (DSAs), still lack exploration. Recently, ChatChisel\cite{Charchisel}, a Chisel-based agile hardware design workflow, successfully designs an RV32I RISC-V CPU with 5-stage pipeline. DSAs, however, involve highly customized and specialized hardware architectures compared to general CPU/GPU, which presents greater difficulty for LLM generation. 

To address this issue, \cite{GPT4AIGCHip} proposes the GPT4AIGChip framework, which innovatively utilizes demo-augmented prompt-generation pipeline to automate template-based AI accelerator design with LLMs. SA-DS\cite{SA-DS} creates a spatial array design dataset. It enables design reuse and customization based on Berkeley's Gemmini\cite{Gemmini} template. However, these works do not fully address the generation challenges of accelerators, as they overlook the generation of the entire hierarchical design, restricting LLMs to fill in only the fixed-level content. ROME\cite{ROME}, on the other hand, proposes an automatic hierarchical generation pipeline with no human feedback. Yet, for each user input, it generates the hierarchical code from scratch, disregarding module reuse. Furthermore, it focuses solely on generation accuracy and ignores the design’s PPA metrics.
\section{Framework}
\label{sec:fram}

\subsection{Overview}

In our HiVeGen framework, as shown in Figure \ref{fig1}, we leverage a text-based natural language description as user inputs. The framework supports two modes of design. We directly generate the hierarchy-aware prompts with Design Space Explorer for simple designs without templates. For application-oriented DSAs, conversely, we employ an LLVM-based kernel extractor to analyze the input application written in C/C++ and extract its corresponding Data Flow Graph (DFG) as well as necessary properties through semantic analysis. The (a) Hierarchy-Aware Prompt Generation Engine then produces the augmented prompts based on kernel information and design templates. The hierarchy-aware prompt will then be sent to the (b) On-the-fly Parsing Engine to facilitate task planning and real-time interaction for users, thereby reducing the cost of error correction. We further develop the (c) Weight-Based Retrieving Engine to enhance module reuses by establishing a dynamically adjusted Code Library. The engine retrieves codes from the library, generates modules based on the task order, and finally assembles all the modules. After that, the generated hierarchical RTL design will go through code validation and PPA evaluation to judge whether the design meets our requirements. If it fails, the context related to PPA optimization will go back to the Design Space Explorer and call the next-round generation.

\subsection{Hierarchy-aware Prompt Generation Engine}
Utilizing the insights that the initial input provided by users may lack the necessary hierarchy details for LLM to generate corresponding designs, we develop a Hierarchy-Aware Prompt Generation Engine in our framework, which involves an LLM-based Design Space Explorer, a Prompt Enhancer, and a Configuration Evaluator. 

Figure \ref{fig1}(a) illustrates the flow of the Hierarchy-Aware Prompt Generation Engine that supports two design modes. For simple designs that do not have pre-defined templates, the Design Space Explorer takes in the natural language description and directly identifies the optimized prompt with hierarchical structure and augmented details utilizing LLM agents based on previous PPA feedback. However, for complicated designs, including highly-specialized DSAs, it is challenging for LLMs to independently explore all hierarchy-based details. To address this, the Design Space Explorer will first perform an optimal design search based on hierarchical templates with configurable parameters, which narrows the design space. The Prompt Enhancer will then proceed with the generated JSON configuration to produce the enhanced prompt. For the two modes of designs, i.e., simple and complicated designs, we provide two distinct PPA-aware system prompts to enable the generation of a direct prompt and a configuration JSON file with explorable parameters, respectively. 

Observing that parameter coupling and interdependency exist in the templated-based design, we developed a Configuration Evaluator to check the design rule conflicts in the generated configuration. If the prompt fails to meet the design rules, it will be fed back to the Design Space Explorer to produce new configurations; otherwise, it will be fed back to the next-stage parser. 

Through this prompt generation flow, we have implemented PPA-aware design space exploration with hierarchy decomposition, which empowers LLMs with hierarchy-aware exploration abilities, and ensures the correctness of configuration based on certain design rules and goals. 

\begin{figure}[h]
     \centering
\includegraphics[width=1.0\linewidth]{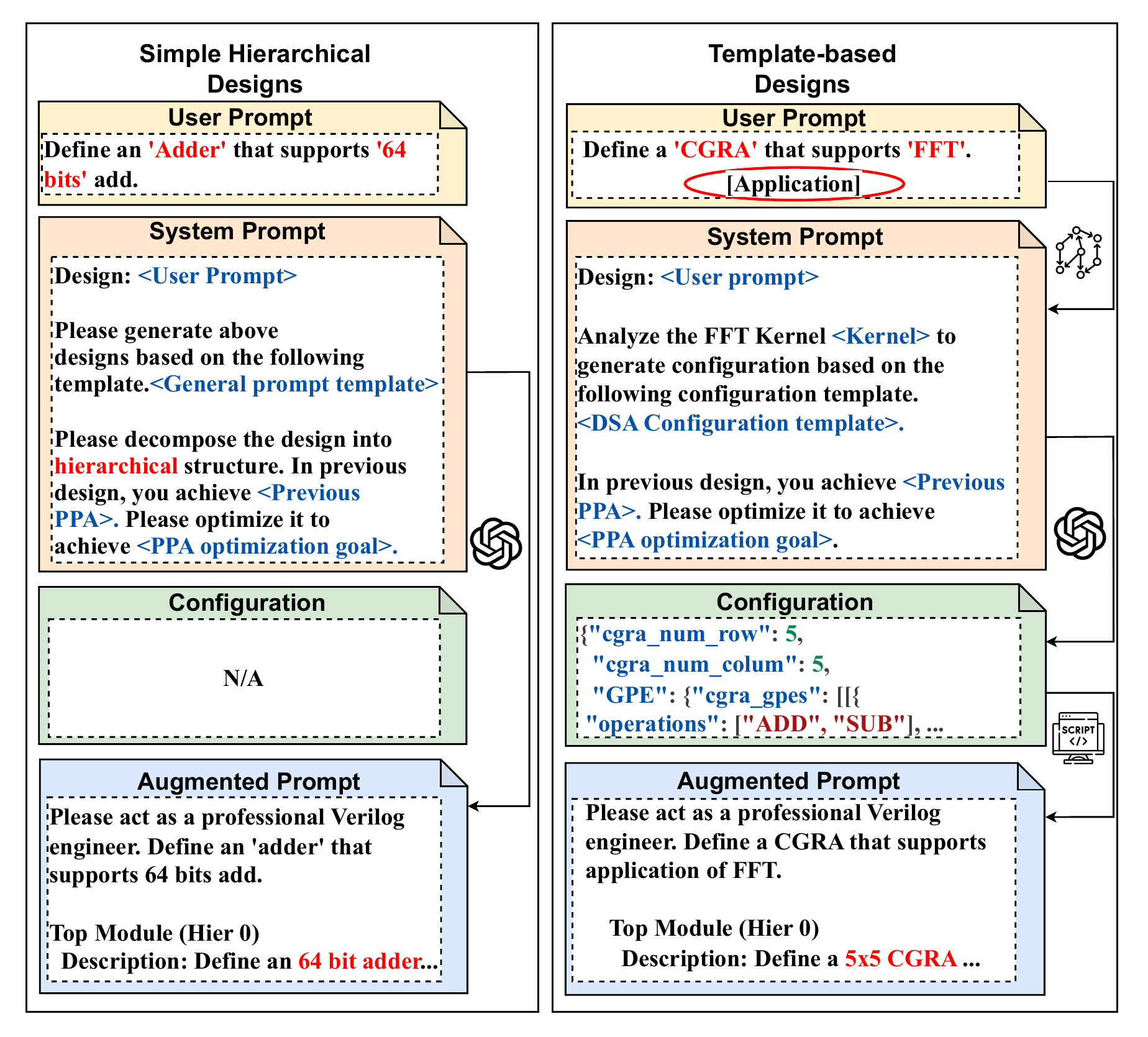} 
    \caption{Prompt examples for two design modes.}
    \label{fig2}
     \vspace{-4pt}
\end{figure}

Figure \ref{fig2} demonstrates the prompt examples for two modes of design. Both the system prompts are PPA-aware with previous-round feedback from the PPA checker, with template-based designs offering additional application kernel properties and templates with parameters that can be configured. Considering that directly generating the augmented prompts with descriptions for the complicated design consumes a substantial amount of tokens, we produce the concise configuration file with LLMs rather than the prompt itself, which also saves I/O cost. Users may also opt to provide suggested PPA optimization strategies to the Design Space Explorer, which enables further performance improvement.

\subsection{On-the-fly Parsing Engine}
The On-the-fly Parsing Engine is composed of two components: a Task Manager and a Runtime Parser, as shown in Figure \ref{fig1}(b).

We first introduce a LLM-based Task Manager that primarily extracts the required modules based on input prompt hierarchy. In other words, it retrieves the module names from the prompt, deduplicates and sorts the generation order of modules. The task manager generates a task list composed of module names, which serves as the reference for subsequent modification and retrieval.

In traditional automation, there is often a lack of interface for users to view or modify the code structures in real time. Consequently, errors such as incorrect port definitions and mismatched module names can only be corrected after generating the entire design, which is costly. By incorporating the proposed Runtime Parser, users can receive immediate feedback on HDL code formulations without accessing outputs while providing modification prompts to the Runtime Parser interactively. 

 \begin{figure*}[h]
     \centering
\includegraphics[width=1.0\linewidth]{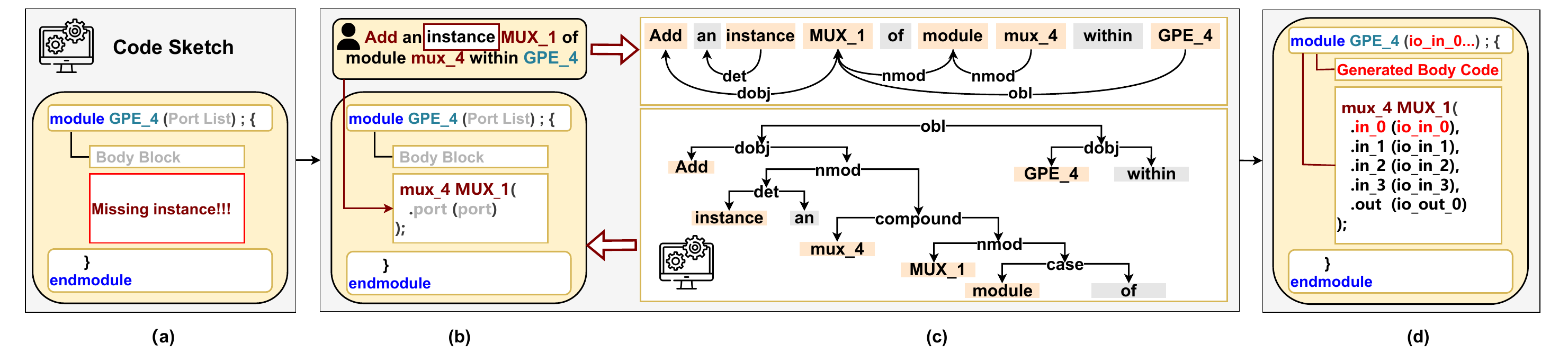} 
    \vspace{-1em}
    \caption{Parser working principle. (a) Initial sketch. (b) User input and refined structure from Parser. (c) Parser working process. (d) Refined code from LLMs. }
    \label{figure:Code_generation1}
\end{figure*}

We use prompt \textit{``Add an instance MUX\_1 of module mux\_4 within GPE\_4"} as an illustration, as shown in Figure~\ref{figure:Code_generation1}. Users first view the initial code sketch generated by the parser and identifies a missing instance within it, as shown in Figure~\ref{figure:Code_generation1}(a). Then they are allowed to input the suggested modification prompt. In Figure~\ref{figure:Code_generation1}(c), the Runtime Parser starts by extracting the linguistic structure of the input sentence, as visualized in a linguistic structure (LS) Tree~\cite{kuebler2009dependency}.
\texttt{Add} serves as the \textsf{root} of the sentence's dependency tree,
and determines the main action or \textsf{verb} of the sentence. 
\texttt{instance} is identified as the direct object (\textsf{dobj}) of the \textsf{verb} "add", 
and becomes the target of the HDL definition.
\texttt {mux\_4} functions as a prepositional modifier in the sentence, 
\texttt{mux\_4} introduces information about how the mux\_4 is constructed, specifically referencing the use of modules in the HDL design.
\texttt{MUX\_1} serves as a noun phrase (\textsf{NP}) that modifies the word ``mux\_4" to describe what type of application will be used.
Figure~\ref{figure:Code_generation1}(b) shows the refined code sketch generated from the Runtime Parser. 
The phrase ``add an instance MUX\_1" is translated into
\texttt{mux\_4 MUX\_1 (.port(port));}. The placeholder \texttt{body block} is the module internal logic, to be implemented. Figure~\ref{figure:Code_generation1}(d) illustrates how the LLM refines the code draft.

By continuously modifying the draft HDL code sketch and the task lists, our Runtime Parsing Engine will largely improve the generation quality without extra token costs.

\subsection{Weight-based Retrieving Engine}
To save generation time and effort by utilizing module reuses, we present a Weight-Based Retrieving Engine composed of a Code Retriever, a Module Generator, and a Prompt Assembler. The Code Library from which codes are retrieved further integrates two essential mechanisms, that are weight management and garbage collection, as illustrated in Figure \ref{fig1}(c).

Each code generation will be on its own thread, thus improving the performance of the code generation. Within the thread, the code retriever looks in the code database based on cosine similarity and weight based on Eq. \ref{eq1}:

\begin{align}
\mathrm{Code}(\mathbf{T}) = \mathrm{Code} \big(\arg\max_{\mathbf{T}_{\mathrm{db}} \in \mathrm{DB}} \big( \cos(\mathbf{T}, \mathbf{T}_{\mathrm{db}}) \cdot w(\mathbf{T}_{\mathrm{db}}) \big) \big)
\label{eq1}
\end{align}
\noindent where $\mathbf{T}$ is the token used to retrieve code, and $\mathbf{T}_{\mathrm{db}}$ is token in the Code Library.

If no such code block reaches the controlling threshold, the LLMs will be called through API to generate a new block with specifications. To further increase the stability and reliability of the Code Library, each code will be automatically verified by the LLM-generated testbench before they enter it. After all code blocks are retrieved and filled in the code sketch, the main function will be generated with the input and output pins of the sub-modules as part of the prompt. The result will subsequently be sent to the Code Validator and PPA Checker for evaluation, and the Code Library will maintain itself through weight management and garbage collection, depending on whether the evaluation succeeds.

\begin{algorithm}
\caption{Dynamic Weight Management for Code Blocks}
\label{alg:dynamic_sampling}
\begin{algorithmic}[1]
\REQUIRE Original weight set to 0.5
\FOR{$n = 1$ to $N$}
    \IF{module $n$ is a part of success}
        \STATE $W_n \mathrel{*}= 1.06$
    \ELSE
        \IF{$W_n < 0.3$ \textbf{and} $second\_chance$}
            \STATE $W_n = 0.5$
            \STATE $second\_chance = False$
        \ELSE
            \STATE $W_n \mathrel{*}= 0.9$
        \ENDIF
    \ENDIF
\ENDFOR
\STATE Start garbage collection on $W_n < 0.2$
\STATE Add each satisfied entry's hash to the avoidance table.
\end{algorithmic}
\end{algorithm}

We design two mechanisms to manage the Code Library. First, in weight management, each code block is assigned a dynamic weight, with implementations with higher success rates given higher weights. The Code Retriever considers both weights and embedded vector similarity when fetching code. Additionally, to prevent blocks with high weights from being overly favored, we implement a ``second chance" mechanism, in which lower-weight blocks will get a chance after their ``siblings" are retrieved \textit{m} times. If a code block still has poor performance after \textit{j} retrievals, it will be marked and fed into the garbage collection process. The garbage collection process, on the other hand, identifies code blocks marked as ready for collection and compares them with the most similar code blocks in the library. This aims to refine the lower-weight code blocks, which is referred to as the ``third chance". 
If all the attempts fail, we remove the code block from the Code Library and record its hash value to avoid later-generated identical code blocks being stored again. The pseudocode for these two mechanisms is shown in Algorithm \ref{alg:dynamic_sampling}.

\section{Evaluation}
\begin{table*}[ht]
    \centering
    \caption{Generation accuracy comparison on maunual-written hierarchical benchmarks over 3 designs.}
    \renewcommand{\arraystretch}{1.2}
    \setlength{\tabcolsep}{3pt}
    \begin{tabular}{c|ccc|ccc|ccc|ccc}
        \hline
        & \multicolumn{3}{c|}{\textbf{64-to-1 Multiplexer}} & \multicolumn{3}{c|}{\textbf{5-to-32 Decoder}} & \multicolumn{3}{c|}{\textbf{32 bit Barrel Shifter}} & \multicolumn{3}{c}{\textbf{UART 8-bit}} \\
        & Pass@1 & Pass@5 & Avg. Time(s) & Pass@1 & Pass@5 & Avg. Time(s) & Pass@1 & Pass@5 & Avg. Time(s) & Pass@1 & Pass@5 & Avg. Time(s)  \\
        \hline
        GPT-3.5 (H) & 0.1 & 0.5 & 7.92 & 0.4 & 0.976 & 8.34 & 0.0 & 0.0 & 7.19 & 0.0 & 0.0 & 6.19\\
        ROME\cite{ROME} & 0.7 & 1.0 & 327.34 & \cellcolor[HTML]{C3E6CB}0.8 & 1.0 & 215.33 & 0.1 & 0.5 & 18.27 & \cellcolor[HTML]{C3E6CB}0.7 & 1.0 & 564.23\\

        HiVeGen (ours) & \cellcolor[HTML]{C3E6CB}0.8 & 1.0 & \cellcolor[HTML]{C3E6CB}5.68 & 0.6 &1.0 & \cellcolor[HTML]{C3E6CB}5.08 & \cellcolor[HTML]{C3E6CB}0.8 & \cellcolor[HTML]{C3E6CB}1.0 & \cellcolor[HTML]{C3E6CB}3.64 & 0.5 & 1.0 & \cellcolor[HTML]{C3E6CB}2.90 \\
        \hline
        GPT-4 (H) & 0.4 & 0.976 & 27.04 & 0.5 & 1.0 & 27.48 & 0.0 & 0.0 &  27.39 & 0.0 &  0.0 & 32.26\\
        ROME\cite{ROME} & 0.9 & 1.0 & 507.54 & 1.0 & 1.0 & 379.65 & 0.7 & 1.0 & 42.44 & \cellcolor[HTML]{C3E6CB}0.8 & 1.0 & 752.76\\

        HiVeGen (ours) & \cellcolor[HTML]{C3E6CB}1.0& 1.0& \cellcolor[HTML]{C3E6CB}6.65& 1.0 & 1.0 & \cellcolor[HTML]{C3E6CB}7.83 &\cellcolor[HTML]{C3E6CB}1.0 &1.0 & \cellcolor[HTML]{C3E6CB}8.28& 0.7 &1.0 &\cellcolor[HTML]{C3E6CB}27.13\\
        \hline
        O1 (H) & 0.8 & 1.0 & 13.63 & 1.0 & 1.0 & 11.70 & 0.1 & 0.5 &  11.16 & 0.0 & 0.0 & 13.67\\
        HiVeGen (ours) & \cellcolor[HTML]{C3E6CB}0.9 & 1.0 & \cellcolor[HTML]{C3E6CB}6.23 & 1.0 &1.0 &  \cellcolor[HTML]{C3E6CB}10.41 & \cellcolor[HTML]{C3E6CB}0.9 & \cellcolor[HTML]{C3E6CB}1.0 & \cellcolor[HTML]{C3E6CB}8.58 & \cellcolor[HTML]{C3E6CB}0.8 & 1.0 &\cellcolor[HTML]{C3E6CB} 13.80\\
        \hline
    \end{tabular}
\label{table1}
\end{table*}
\subsection{Experimental Setup}
\begin{figure}[h!]
     \centering
\includegraphics[width=1.0\linewidth]{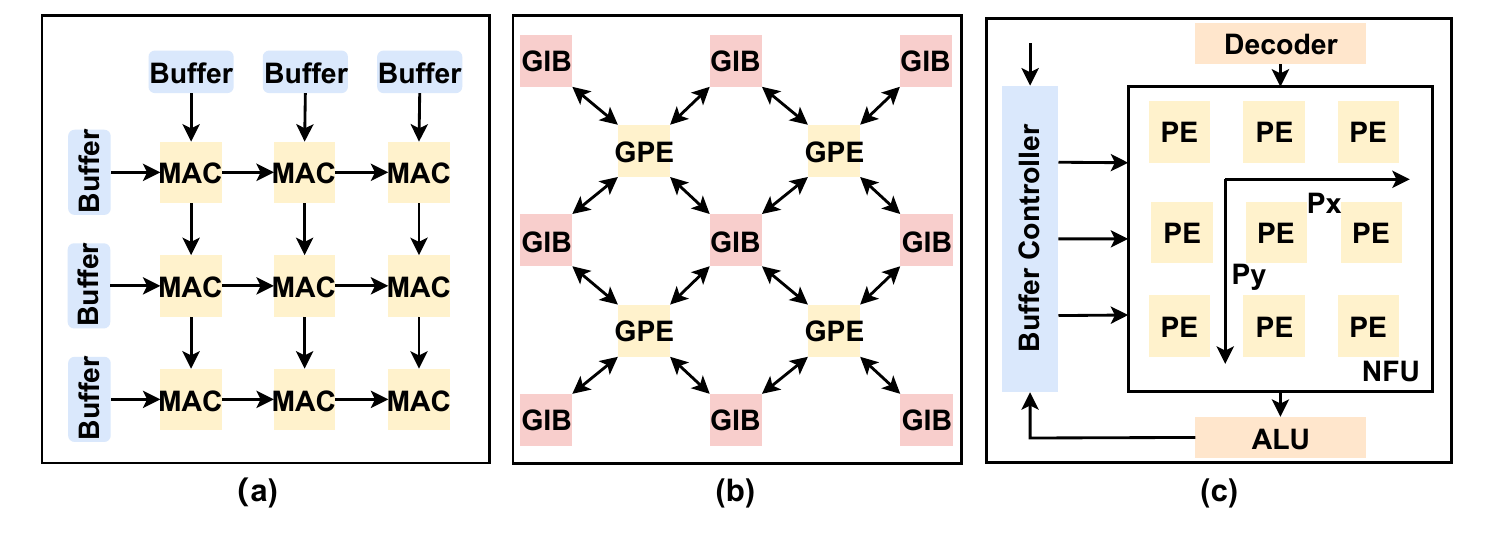} 
    \caption{Domain-Specific Accelerators examples. (a) Systolic Array. (b) CGRA. (c) ShiDianNao. }
    \label{fig4}
\end{figure}

\textbf{Dataset.} To ensure fair comparisons with existing work, our experiment adopts two datasets. The first dataset is composed of simple designs to be generated with Path \textcircled{2} in Figure \ref{fig1}, including the Multiplexer, the Decoder, the Barrel Shifter, and the UART. These designs are the same as the first dataset in ROME\cite{ROME}. Given the focus of this study on DSAs, the second dataset selects and evaluates three hierarchical DSA designs for different reasons, which involves a commonly-used Systolic Array with buffers, a renowned Convolutional Neural Network accelerator ShiDianNao\cite{shidiannao}, and Coarse-grained Reconfigurable Architecture (CGRA)~\cite{fdra} with a relatively larger design space, as shown in Figure \ref{fig4}. Note that our generation of CGRA only involves the GPE and GIB array, and the generation of ShiDianNao only involves the NFU, which is a PE array.

\textbf{Configuration and Platforms.} Our HiVeGen employs three commercial LLM models to generate RTL hierarchy-aware designs, that are GPT-3.5-Turbo, GPT-4, and O1-mini. We use top-p = 0.9 and temperature = 0.5 as the basic configuration. For syntax and functional verification, we integrate the ICARUS Verilog simulator\cite{iverilog} to automate the workflow. We also synthesize the LLM-aided hardware design based on a TSMC 28nm technology using Design Compiler. All experiments run on a Windows-Sub-Linux CPU server with Intel(R) i9-13900KF and Nvidia GeForce RTX 4090. In our data collection, our max trying time k=3, second chance trigger point m = 10, garbage marking j = 30.

\textbf{Experimental Metric.}  We evaluate the experimental results using multiple metrics. For generation accuracy, we apply Pass@1 and Pass@5 metrics to assess the generation pass rate. Additionally, the average generation time and token savings are measured, thus illustrating the design speed improvement and cost savings through our framework. We also evaluate the generation quality, i.e., PPA optimization, based on metrics including power, clock and area.
\begin{table}[h!]
    \centering
    \caption{Generation accuracy and token savings comparison. Generation accuracy is evaluated on O1, while token savings is measured on GPT-4.}
    \renewcommand{\arraystretch}{1.4}
    \setlength{\tabcolsep}{1pt}
    \textbf{(a) Generation accuracy comparison using O1 over three designs.}
    \vspace{0.2cm}
    
    \begin{tabular}{c|cc|cc}
        \hline
        & \multicolumn{2}{c|}{\textbf{ROME\cite{ROME}}} & \multicolumn{2}{c}{\textbf{HiVeGen (ours)}} \\

        &  Pass@1 & Pass@5 &Pass@1 & Pass@5  \\
        \hline
        Multiplexer & 0.5 & 0.996 & \cellcolor[HTML]{C3E6CB}1.0 & \cellcolor[HTML]{C3E6CB}1.0  \\
        \hline
        Decoder & 1.0 & 1.0 & 1.0 & 1.0 \\
        \hline
        Barrel Shifter & 0.3 & 0.917 & \cellcolor[HTML]{C3E6CB}1.0 & \cellcolor[HTML]{C3E6CB}1.0 \\
        \hline
    \end{tabular}
    
    \vspace{0.5cm}
    \textbf{(b) Token savings compared to directly using GPT-4 over Multiplexer.}

    \vspace{0.2cm}
    
    \begin{tabular}{c|c|c|c}
        \hline
        & \multicolumn{2}{c|}{\textbf{Tokens}} & \textbf{Token Savings (\%)} \\
        \cline{2-3}
        \textbf{} & \textbf{GPT-4 (NH)} & \textbf{HiVeGen (ours)} & \\
        \hline
        Multiplexer & 2089 & \cellcolor[HTML]{C3E6CB}1442 & 30.97 \\
        \hline
    \end{tabular}

\label{table2}
\end{table}

\subsection{Generation Accuracy}

\begin{table*}[h!]
    \centering
    \caption{PPA optimization result including Pass@5, DSE defined scale, power (mW), clock (ns) and area (um\textsuperscript{2}). The whole HiVeGen framework generates Systolic array, CGRA and ShiDianNao for GEMM, FFT, GEMM application respectively.}
    \renewcommand{\arraystretch}{1.2}
    \setlength{\tabcolsep}{4pt}
    \begin{tabular}{c|ccccc|ccccc|ccccc}
        \hline
        & \multicolumn{5}{c|}{\textbf{Systolic Array (GEMM)}} & \multicolumn{5}{c|}{\textbf{CGRA (FFT)}} & \multicolumn{5}{c}{\textbf{ShiDianNao (GEMM)}}\\

        &  Pass@5 & Scale & Power & Clock & Area & Pass@5 & Scale & Power & Clock & Area & Pass@5 & Scale & Power & Clock & Area   \\
        \hline
        GPT-3.5 & 0 & / & / & / & / & 0 & / & / & / & /  & 0 & / & / & / & /  \\
        HiVeGen (w/o ICL) & 0 & / & / & / & / & 0 & / & / & / & / & 0 & / & / & / & / \\
        HiVeGen (w/ ICL) & 0 & / & / & / & / & 0.25 & 5$\times$5 & 13.5 & 0.11 &  54064 & 0 & / & / & / &/ \\
        \hline
        GPT-4 & 0 & / & / & / & / & 0 & / & / & / & /  & 0 & / & / & / & /\\
        HiVeGen (w/o ICL) & 0 & / & / & / & / & 0.25 & 8$\times$8 & 34.6 & 0.22 &138400 & 0 & / & / & / & / \\
        HiVeGen (w/ ICL) & 0.25 & 5$\times$5 & 9.63 & 4.6 & \cellcolor[HTML]{C3E6CB}27022 & 0.45 & 2$\times$2 & 13.8 & 0.05 & 8650 & 0 & / & / & / & / \\
        \hline
        O1 & 0 & / & / & / & / & 0 & / & / & / & /  & 0 & / & / & / & / \\
        HiVeGen (w/o ICL) & 0.25 & 5$\times$5 & 9.33 & 4.6 & 27286 & 0 & / & / & / & / & 0 & / & / & / & / \\
        HiVeGen (w/ ICL) & \cellcolor[HTML]{C3E6CB}0.45 & 5$\times$5 & \cellcolor[HTML]{C3E6CB}8.7 & \cellcolor[HTML]{C3E6CB}0.05 & 30168  & \cellcolor[HTML]{C3E6CB}0.60 & 2$\times$2 & \cellcolor[HTML]{C3E6CB}5.5 & 0.05 & \cellcolor[HTML]{C3E6CB}5196 & \cellcolor[HTML]{C3E6CB}0.25 & 5$\times$5 & \cellcolor[HTML]{C3E6CB}19.12 & \cellcolor[HTML]{C3E6CB}0.5 & \cellcolor[HTML]{C3E6CB}69595 \\
        \hline
    \end{tabular}
\label{table3}
\end{table*}
\vspace{0.5cm}

\textbf{Manual-written hierarchical prompt.} To enable a fair comparison of generation accuracy and speed, our experiment initially employs a manual-written hierarchical prompt directly using Engine (b) and (c) in Figure \ref{fig1} with no user input modification. We allow a total of 10 generations. As illustrated in Table \ref{table1}, the performance of our generation accuracy increases impressively compared to GPT with hierarchical prompt, and also ROME, especially in design Barrel Shifter. This results from more detailed sub-task division and quality-assured submodule retrieval facilitated in our HiVeGen framework. Additionally, the generation time has been significantly reduced by 45.24\% and by 97.61\% on average compared to the pure GPT and ROME respectively, which can be attributed to the implementation of threading within the framework and the module reuse.

\textbf{LLM-generated hierarchical prompt.} To further demonstrate the advantages of our Hierarchy-aware Prompt Generation Engine, we incorporate Engine (a) and real-time error correction mechanism in Figure \ref{fig1} into our experiments to provide a fair comparison with ROME on the LLM-based automatic generation of the hierarchical prompt. The results indicate that Engine (a) and the Runtime Parser lead to a slight improvement in generation accuracy due to better consistency and real-time error correcting interaction using O1, particularly for Barrel Shifter and Multiplexer, with Pass@1 increasing from 0.9 to 1.0. When compared to ROME, our generation's accuracy is impressively superior as well. Additionally, we measure the proportion of token savings when generating a Multiplexer compared to directly using GPT-4 with non-hierarchical prompt. Due to the hierarchical design and concise prompts, our token usage is reduced by 30.97\%, which means that our HiVeGen framework serves as an effective solution to the token limit.

\subsection{PPA Optimization}

For Domain-Specific Accelerators shown in Figure \ref{fig4}, we utilize Path \textcircled{1} in Figure \ref{fig1} to generate our design. We allow a total of 20 generations for each design, with four correction attempts within each of the 5 generation iterations. Our generation pipeline is application-oriented, which means that LLM will automatically explore the best configuration for the specified application. We employ user input \textit{``Define a CGRA for application FFT"} as an illustration. The kernel extractor will first analyze the kernel FFT and generate the DFG with operator information, which will then be fed to the LLM-based Design Space Explorer to define the configuration of CGRA. For example, LLM agent defines the scale of CGRA that supports the FFT kernel to be 2$\times$2, and the ALU supported operations in GPE to be ``PASS, ADD, SUB". 

We divide HiVeGen-based experiments into two groups. The first group generates without utilizing a PPA-aware in-context learning (ICL) template, which means that it will generate the configuration with no shot and only a blank template, whereas the second group incorporates the PPA-emphasized one-shot JSON template into prompt generation, with a concise text file involving parameter explanation, PPA feedback, objectives, and suggested optimization strategies. Here, we position ``pipelining" as the suggested strategy and ``clock" as the objective. 

Our experiments display that utilizing HiVeGen enhances the generation accuracy compared to pure LLMs. Specifically, Pass@5 rises from 0 to 0.25 for Systolic Array with O1 and 0 to 0.25 for CGRA with GPT-4.

When integrating the PPA-aware ICL template, the LLMs will further optimize the design based on users' proposed objective while exhibiting an improved success rate due to the one-shot template. Insightfully, we have the following three observations:
(1) the configuration decisions made by GPT-3.5 during the generation with ICL templates are heavily influenced by the shots provided, resulting in suboptimal outcomes;
(2) the proposed ``pipelining" strategy suggests HiVeGen to incorporate registers into the design, thereby achieving improved clock frequency with ICL template;
(3) despite the same scale of CGRA, O1 outperforms GPT-4, attributed to its simplified ALU design, which is also a configurable parameter in the template. This, to some extent, highlights the flexibility of the O1 model. To summarize, our HiVeGen framework demonstrates significant potential in more complicated accelerator designs with larger design space due to its impressive performance in generating above DSAs. 

\label{sec:eval}
\vspace{0.5cm}
\section{Conclusion}
\label{sec:conclusion}

In this paper, we propose HiVeGen, a hierarchical LLM-based automatic Verilog generation framework. HiViGen exhibits its potential to be an effective solution to token limits, code redundancy, and error correction costs when generating complex designs with LLMs. To the best of our knowledge, we are the first to propose a hierarchy-aware hardware accelerator generation framework.  Extensive experiments show that HiVeGen can reach up to 30.97\% token savings and 45.24\% time savings on average compared to pure GPT generation with an improved generation accuracy. Our framework also supports PPA-aware DSA generation, which contributes to the advanced quality of LLM-generated design.

\bibliographystyle{plain}
\bibliography{references, ding}

\end{document}